\documentclass[sn-mathphys-num]{sn-jnl}

\usepackage{graphicx}%
\usepackage{multirow}%
\usepackage{amsmath,amssymb,amsfonts}%
\usepackage{amsthm}%
\usepackage{mathrsfs}%
\usepackage[title]{appendix}%
\usepackage{xcolor}%
\usepackage{textcomp}%
\usepackage{manyfoot}%
\usepackage{booktabs}%
\usepackage{algorithm}%
\usepackage{algorithmicx}%
\usepackage{algpseudocode}%
\usepackage{listings}%
\usepackage{float}%
\usepackage{mathptmx}
\usepackage{hyperref}
\renewcommand{\normalsize}{\fontsize{12}{14.4}\selectfont}

\usepackage{titlesec}
\titleformat{\section}
  {\normalfont\fontsize{14}{17}\bfseries}{\thesection}{1em}{}

\usepackage{lineno}   

\usepackage{caption}
\usepackage[T1]{fontenc} 
\DeclareCaptionLabelSeparator{pipe}{ | }  
\newcommand{\comment}[1]{}
\theoremstyle{thmstyleone}%
\theoremstyle{thmstyletwo}%

\theoremstyle{thmstylethree}%

\begin{document}
\nolinenumbers
\title[Article Title]{\textbf{pUniFind: a unified large pre-trained deep learning model pushing the limit of mass spectra interpretation}}


\author[1,3]{\fnm{Jiale} \sur{Zhao}}\email{zhaojiale22z@ict.ac.cn}

\author[1,3]{\fnm{Pengzhi} \sur{Mao}}
\author[1,3]{\fnm{Kaifei} \sur{Wang}}
\author[3]{\fnm{Yiming} \sur{Li}}
\author[1,3]{\fnm{Yaping} \sur{Peng}}
\author[1,3]{\fnm{Ranfei} \sur{Chen}}
\author[2]{\fnm{Shuqi} \sur{Lu}}
\author[2]{\fnm{Xiaohong} \sur{Ji}}
\author[1,3]{\fnm{Jiaxiang} \sur{Ding}}
\author[3]{\fnm{Xin} \sur{Zhang}}
\author[5]{\fnm{Yucheng} \sur{Liao}}
\author[4,5,6]{\fnm{Weinan} \sur{E}}
\author*[2]{\fnm{Weijie} \sur{Zhang}}\email{zhangweijie@dp.tech}
\author*[2,4,7]{\fnm{Han} \sur{Wen}}\email{wenh@dp.tech}
\author*[1,3]{\fnm{Hao} \sur{Chi}}\email{chihao@ict.ac.cn}

\affil[1]{\orgdiv{Key Laboratory of Intelligent Information Processing of Chinese Academy of Sciences (CAS)}, \orgname{Institute of Computing Technology}, \orgaddress{\city{Beijing}, \postcode{100190}, \state{Beijing}, \country{China}}}

\affil[2]{\orgdiv{DP Technology Co., Ltd.} \orgaddress{\city{Beijing}, \country{China}}}

\affil[3]{\orgdiv{University of Chinese Academy of Sciences}, \orgaddress{\city{Beijing}, \country{China}}}

\affil[4]{\orgdiv{AI for Science Institute}, \orgaddress{\city{Beijing}, \country{China}}}

\affil[5]{\orgdiv{Center for Machine Learning Research}, \orgname{Peking University}, \orgaddress{\city{Beijing}, \country{China}}}

\affil[6]{\orgdiv{School of Mathematical Sciences}, \orgname{Peking University}, \orgaddress{\city{Beijing}, \country{China}}}

\affil[7]{\orgdiv{State Key Laboratory of Medical Proteomics}, \orgaddress{\city{Beijing 102206}, \country{China}}}


\abstract{
Deep learning has advanced mass spectrometry data interpretation, yet most models remain feature extractors rather than unified scoring frameworks. We present pUniFind, the first large-scale multimodal pre-trained model in proteomics that integrates end-to-end peptide-spectrum scoring with open, zero-shot de novo sequencing. Trained on over 100 million open search-derived spectra, pUniFind aligns spectral and peptide modalities via cross-modality prediction and outperforms traditional engines across diverse datasets, particularly achieving a 42.6\% increase in the number of identified peptides in immunopeptidomics. Supporting over 1,300 modifications, pUniFind identifies 60\% more PSMs than existing de novo methods despite a 300-fold larger search space. A deep learning–based quality control module further recovers 38.5\% additional peptides—including 1,891 mapped to the genome but absent from reference proteomes—while preserving full fragment ion coverage. These results establish a unified, scalable deep learning framework for proteomic analysis, offering improved sensitivity, modification coverage, and interpretability.

}




\maketitle

\section*{Introduction}\label{sec1}

Proteomics has become a pivotal field in modern biological research, with tandem mass spectrometry (MS/MS) established as a foundational analytical technique\cite{aebersold2016mass}. Database search engines serve as the primary tools for peptide and protein identification from MS/MS data, relying on pre-existing sequence databases to match experimental spectra to theoretical peptide fragments. Traditional search engines, such as SEQUEST\cite{eng1994approach}, MaxQuant\cite{cox2008maxquant}, pFind\cite{chi2018comprehensive} and Alphapept\cite{strauss2024alphapept}, typically employ simple machine learning models to assess peptide-spectrum matches (PSMs) based on a limited set of manually curated features. To further enhance the identification rate of MS/MS data, open search engines such as Open-pFind\cite{chi2018comprehensive} and MSFragger\cite{kong2017msfragger} have been developed to detect peptides resulting from unexpected enzymatic digestions or modifications. In addition, de novo sequencing provides an alternative approach for MS/MS data interpretation by directly inferring peptides without reference to any proteome databases, and plays a pivotal role in applications such as monoclonal antibody sequence assembly and neoantigen discovery. Over the past two decades, several de novo sequencing tools have been introduced, including pNovo\cite{yang2019pnovo}, PEAKS\cite{ma2003peaks}, and PepNovo\cite{frank2005pepnovo}, yet the overall accuracy and robustness remain limited, particularly in complex biological samples\cite{yilmaz2022novo}. 

Despite rapid advancements in both conventional database search and de novo sequencing methods, further improvements in performance remain challenging due to the inherent limitations in their PSM scoring capabilities. Consequently, deep learning has been extensively applied to proteomics data analysis in recent years\cite{mann2021artificial}. Researchers have developed various models to predict key peptide properties, such as the theoretical spectrum patterns\cite{zhou2017pdeep,gessulat2019prosit} and the retention times\cite{ma2018deeprt}, which can be further integrated into machine learning models to enhance PSM scoring\cite{zhou2022pvalid,zeng2022alphapeptdeep,yang2023msbooster}. Recently, Yu et al. proposed DeepSearch, which highlights the potential of end-to-end PSM scoring, and has achieved performance comparable to traditional search engines like MaxQuant\cite{yu2025towards}. Moreover, deep learning techniques have increasingly been applied to address the challenges of de novo sequencing. Following the introduction of DeepNovo\cite{tran2017novo}, several models—such as PointNovo\cite{qiao2021computationally}, Casanovo\cite{yilmaz2022novo,yilmaz2024sequence}, GraphNovo\cite{mao2023mitigating}, and $\pi$-PrimeNovo\cite{zhang2025pi}—have been proposed to improve de novo sequencing performance.


Although deep learning has been widely applied across various aspects of proteomics, its advantages have yet to be fully realized in addressing the core task of computational proteomics—namely, the more sensitive and accurate interpretation of mass spectrometry data for protein and peptide identification. First, to the best of our knowledge, the datasets currently used to train deep learning models in proteomics are primarily derived from conventional database searches, which assume fully tryptic digestion (with rare exceptions) and typically include only a limited set of common modifications. As a result, these datasets do not fully represent the overall characteristics of mass spectrometry data, particularly in the case of non-specific proteolytic digestion or unexpected peptide modifications, which significantly hinder the effective identification of such peptides\cite{chi2018comprehensive}. Consequently, it is crucial to build larger annotated datasets, especially those based on open database search engines, to obtain more comprehensive data for model training. Second, it is well-established that database searching and de novo sequencing share inherent unification in spectrum analysis. However, few studies have proposed unified frameworks that jointly address PSM evaluation and quality control in both database search and de novo sequencing. Although such frameworks are still rare in proteomics, related advances in other fields—particularly the widespread application of multi-task and multimodal learning in computer vision\cite{li2022blip,zhao2021multidimensional}, protein design\cite{hayes2024simulating}, protein function prediction\cite{zhuangpre}, and small molecule pre-training\cite{zhou2023uni}—have demonstrated strong potential to improve generalization, reduce overfitting, and enhance performance across diverse data types. Therefore, jointly considering database search, de novo sequencing, and other critical sub-tasks and designing a unified multimodal model for PSM scoring, is expected to further improve the performance of MS/MS data interpretation.

To address this gap, we present pUniFind, to our knowledge, the first large-scale multimodal pre-trained model in proteomics to systematically address the core challenge of PSM scoring across both database search and de novo sequencing in proteomics data analysis. We employed Open-pFind to annotate a large-scale MS/MS dataset via open database search, encompassing 100 million PSMs from 6,524 MS/MS data files, and adopted a cross-modality prediction training strategy to deepen the model’s understanding of the relationships between mass spectra and peptide features.  The pUniFind-powered database search consistently improves identification performance across diverse species and instruments, notably achieving a 42.6\% increase in peptide yields in immunopeptidomics. More importantly, we comprehensively evaluated the model’s robustness and accuracy using several validation approaches including metabolic labeling, entrapment database search, and mixed-species analysis. In particular, enabled by training data derived from open searches, pUniFind is the first deep learning-based open de novo sequencing method capable of handling over 1,300 distinct modification types. pUniFind identified 60\% more PSMs than conventional de novo methods across datasets enriched with diverse peptide modifications, while achieving performance comparable to that of regular peptides\cite{yilmaz2024sequence}—even within a search space 300 times larger than those used by existing deep learning approaches\cite{yang2017open}. To further
address the lack of reliable quality control mechanisms in conventional de novo sequencing methods\cite{tran2024novoboard}, we developed a novel filtering strategy that integrates multiple features to systematically eliminate unreliable results from diverse perspectives. Using this strategy, our open de novo approach identified 38\% more peptides on immunopeptidomics data than Open-pFind, demonstrating its superior performance.
Notably, 89.3\% of identified peptides were successfully mapped to the human genome, and only 11.6\% lacked detectable binding affinity to human HLA molecules, providing strong evidence for their authenticity. In summary, pUniFind provides a more comprehensive and accurate interpretation of proteomic data, highlighting the potential of end-to-end deep learning approaches to replace conventional scoring methods and delivering more reliable results for downstream applications in life sciences.

\section*{Results}
\textbf{The pUniFind model and its integration into the database search workflow}

\begin{figure}[H]
\centering
\includegraphics[width=0.9\textwidth]{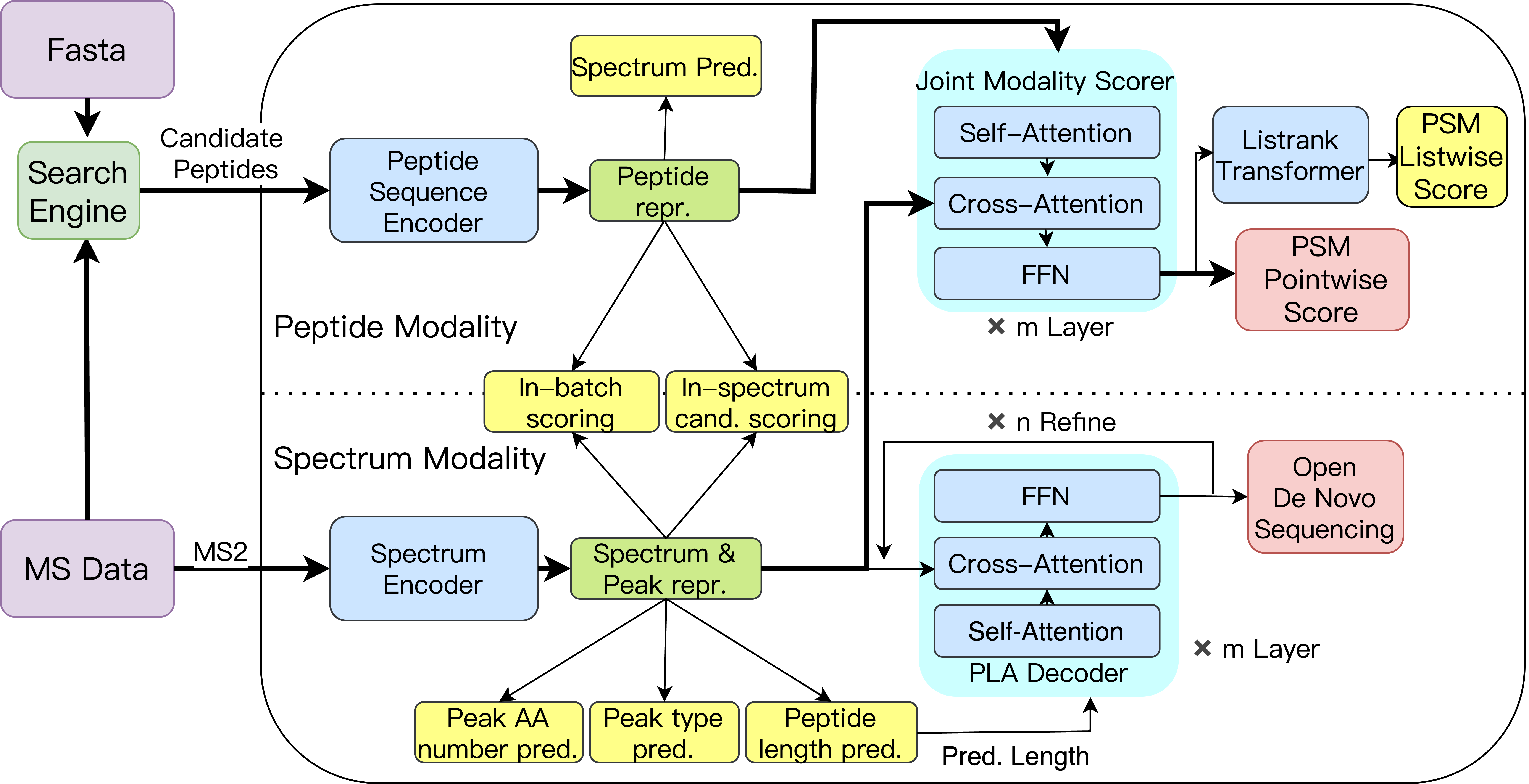}
\caption{\textbf{Architecture and training tasks of pUniFind model.} pUniFind takes peptide sequences and MS/MS spectra as input, which are encoded by dedicated peptide and spectrum encoders to generate modality-specific representations. For each spectrum peak, the corresponding representation is used to predict the number of amino acids in the associated fragment and the fragment ion type (such as b- or y-ion), while the classification head of the spectrum representation predicts the peptide length corresponding to the spectrum. The peptide representation is then used to predict the spectrum. Both peptide and spectrum representations are further utilized for in-batch and in-spectrum scoring. This process segregates peptides from different spectra within a batch and ranks them, selecting negative training candidates from those ranked third to tenth among the top 10 peptide candidates generated by Open-pFind. For the de novo sequencing task, the spectrum representation is passed to the Peptide Length Aware (PLA) module, which predicts the peptide sequence conditioned on the predicted peptide length. Additionally, a joint modality scorer is employed to jointly encode peptide and spectrum embeddings, thereby enhancing the final score used for reranking. We also adopted a listwise ranking task to further improve the representation learning of the joint modality scorer. Bold arrows indicate steps in the database search workflow.
}\label{fig1}
\end{figure}

\textbf{Fig. \ref{fig1}} illustrates the architecture of the pUniFind pre-trained model. The input spectra and peptides are first encoded independently, and their representations are used in several pre-training tasks, as highlighted by the yellow modules in \textbf{Fig. \ref{fig1}}. For database search, these representations are used to predict the pointwise PSM score via a joint modality scorer. To train the model, we reanalyzed large-scale datasets with Open-pFind, collecting over 100 million PSMs. Details of the training procedure and search workflow are provided in the Methods section.

Specifically, to enhance the model’s ability to capture peptide and spectral features, and to improve cross-modal alignment, we employed several pre-training tasks. For peptide representation, we implemented a spectrum prediction task and a candidate ranking task, where transformer layers were used to model inter-candidate relationships within each spectrum. Notably, although listwise scoring performs suboptimally in cross-spectrum ranking, it still proves beneficial when incorporated as an auxiliary task. This improvement is likely driven by joint optimization, which enhances the discriminative capacity of the cross-modal scorer. For MS/MS data, we designed three pre-training tasks: 1) Predicting the number of amino acids for each peak; 2) Predicting the peptide length corresponding to each spectrum; and 3) Predicting whether a specific peak corresponds to b or y ions, as well as neutral losses such as dehydration or ammonia loss.

We also developed a new database search workflow based on pUniFind. Open-pFind serves as the search engine kernel to retrieve the top $k$ candidate peptides for each spectrum—10 by default, or 20 for non-tryptic cases. If the top-ranked candidate has a q-value lower than 0.1, indicating high confidence, all candidates for that spectrum are rescored and reranked using pUniFind. Final identifications are subjected to target-decoy analysis to estimate and control the false discovery rate (FDR). By leveraging Open-pFind to efficiently pre-filter candidates, this workflow significantly reduces the computational burden of pUniFind, enabling large-scale data analysis without the need for exhaustive scoring of unreliable spectra or for evaluating numerous candidate peptides.

\textbf{Performance evaluation on MS/MS data from various species} 

\begin{figure}[H]
\centering
\includegraphics[width=1\textwidth]{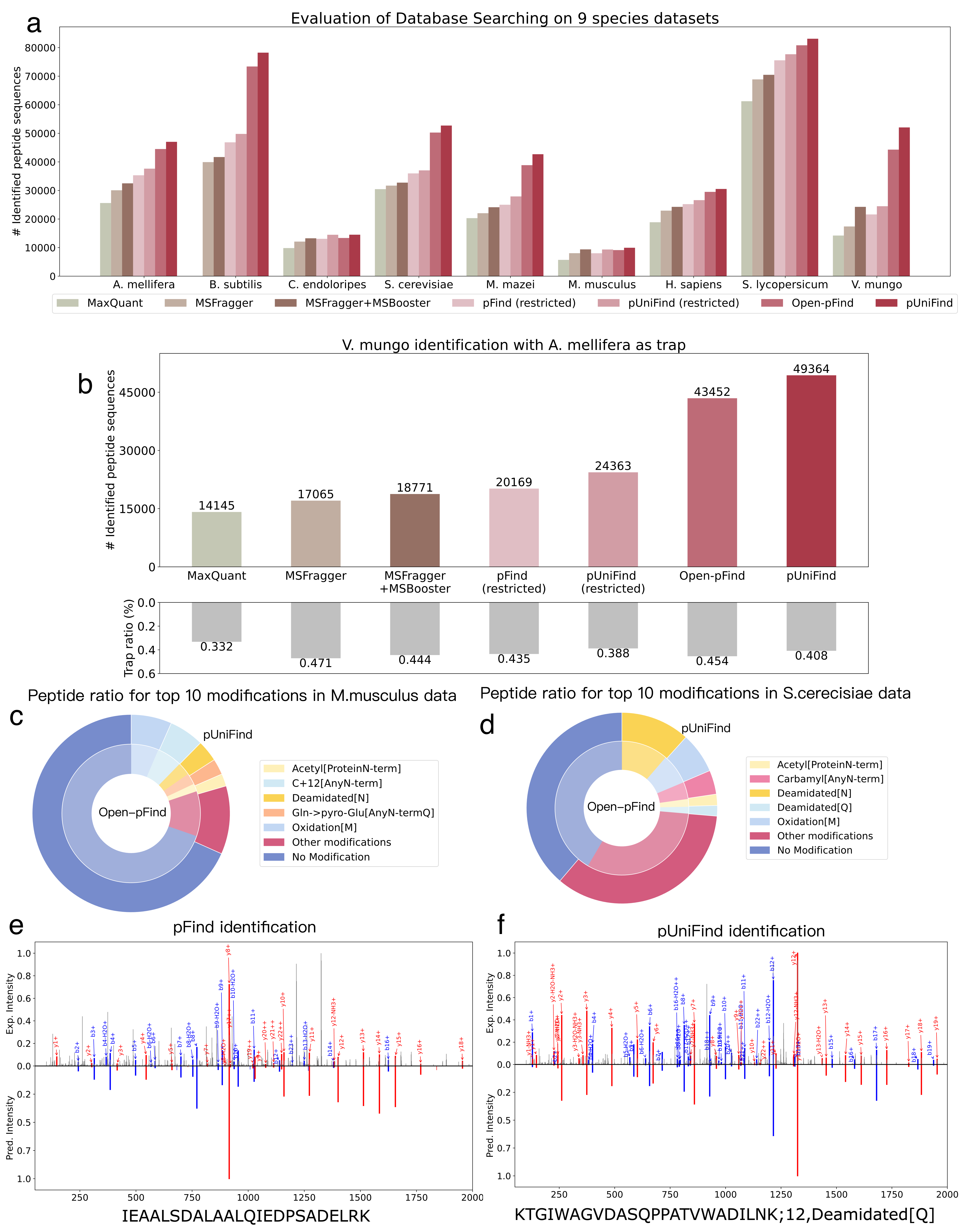}
\caption{
\textbf{Evaluation across multiple species and validation of identification accuracy.} 
\textbf{a.} Performance comparison of various software tools on datasets from nine species, sourced from different laboratories.
\textbf{b.} Entrapment evaluation on \textit{V. mungo} data with \textit{A. mellifera} as the entrapment; neither species was included in the training set.
\textbf{c.} and \textbf{d.} Peptide modification ratios identified by Open-pFind and pUniFind for \textit{M. musculus} and \textit{S. cerevisiae} data respectively.
\textbf{e.} and \textbf{f.} Comparison of predicted and experimental spectra for the same experimental spectrum, using peptide identifications from Open-pFind (unmodified) and pUniFind (deamidated at glutamine, Q); the prediction based on pUniFind shows better agreement with the experimental spectrum.
}\label{fig2}
\end{figure}

We first conducted a comprehensive evaluation of several tools across nine species datasets, encompassing samples from multiple laboratories. Across all these datasets, the open search mode of pUniFind consistently identified the highest number of peptide identifications (\textbf{Fig. \ref{fig2}a}). This advantage can be partially attributed to the open search mode itself—as evidenced by Open-pFind also yielding significantly increased results; however, 
pUniFind outperforms Open-pFind by 2–18\% across peptide-level evaluations and also achieves consistent, albeit slight, improvements at the protein level (\textbf{Supplementary Fig. S1}), highlighting the advantages of its model architecture. In contrast, under the
restricted search mode considering only tryptic peptides with a few common modifications, all software tools, including pUniFind which refines pFind’s results, reported comparable numbers of peptide identifications. While pUniFind maintained a performance edge on most datasets, MSFragger—particularly when enhanced with MSBooster—also achieved a high number of identifications.  Furthermore, an ablation study showed that including the two key cross-modality training tasks, de novo sequencing and spectrum prediction, improved overall performance, with peptide identification increasing by 60\% on \textit{B. subtilis} data (\textbf{Supplementary Fig. S2}). This confirmed that cross-modality training enabled the model to learn more meaningful spectrum and peptide representations while improving alignment and understanding between these two modalities, further validating our pre-training strategy's importance and effectiveness.

The stable performance gains across these nine datasets—particularly the substantial increases in \textit{V. mungo} and \textit{B. subtilis} data—raised concerns about accuracy. To further assess the reliability of the reported results, we applied an entrapment strategy by incorporating the \textit{A. mellifera} proteome as the entrapment database in the analysis of \textit{V. mungo} data analysis (\textbf{Fig. \ref{fig2}b}). Importantly, neither species was present in training data. Given that the ratio of peptides between the \textit{A. mellifera} and \textit{V. mungo} databases is approximately 1:1.75, the estimated expected entrapment peptide ratio at 1\% FDR should be around 0.36\%, calculated as 1 / (1 + 1.75). Remarkably, while pUniFind identified 163.0\% and 144.8\% more peptides than MSFragger with MSBooster and pFind, respectively, and 13.6\% more than Open-pFind, it still maintained a normal or even slightly lower entrapment peptide ratios. This suggests that pUniFind effectively balanced peptide identification performance with accurate entrapment ratios, further validating its robustness and reliability.

We also benchmarked a few other end-to-end PSM scoring frameworks \textit{i.e.} Tesorai\cite{burq2024tesorai} and DDA-BERT\cite{jun2024dda} (\textbf{Supplementary Fig. S3}). Both tools, which lack open search capability, identified significantly fewer peptides than pUniFind, although Tesorai slightly outperformed conventional search engines. Notably, both models exhibited significantly elevated entrapment peptide ratios, likely due to inadvertent memorization of the sequence labels (target or decoy) during pre-training or fine-tuning. This potential label leakage biased the scoring toward targets and inflated false discovery rates. In contrast, our architecture adopted spectrum prediction and de novo sequencing as the primary pre-training objectives, combined with a joint-modality scoring module that is decoupled from the sequence decoder. This design explicitly separates representation learning from scoring, thereby preventing memorization and ensuring robust generalization.

As the degree of improvement in peptide identifications varied substantially across datasets, we further compared the number of modified peptides identified by pUniFind versus Open-pFind for the two model organisms—\textit{M. musculus} and \textit{S. cerevisiae} (\textbf{Fig. \ref{fig2}c-d}). Modified peptides accounted for 30.3\% and 58.4\% of total identifications, in the two datasets, respectively, aligning with the greater  performance gain on \textit{S. cerevisiae} data (\textbf{Fig. \ref{fig2}a}). The consistent modified peptide ratios observed between pUniFind and Open-pFind further validate pUniFind’s robustness in modified peptide scoring, confirming that the expanded search space does not lead to excessive false identifications arising from rare modifications. Traditional scoring methods like Open-pFind penalize modified peptides based on their lower frequency in initial SVM searches, potentially hindering accurate identification. In contrast, pUniFind better utilizes intensity information through its spectrum prediction task and operates as a zero-shot scoring model, avoiding such data bias. \textbf{Fig. \ref{fig2}e} shows poor correlation between predicted and experimental intensities for an unmodified peptide identified by Open-pFind, whereas \textbf{Fig. \ref{fig2}f} shows strong correlation for pUniFind's result for the same spectrum. The consistency analysis of the peptides with the ten most abundant modification types highlights pUniFind’s capability in exploiting low-frequency modifications (\textbf{Supplementary Fig. S4}). Furthermore, we evaluated pUniFind on a benchmark dataset consisting of PSMs annotated with 21 diverse post-translational modifications (PTMs), hereafter referred to as the 21PTM dataset. Despite the absence of percolator-like online refinement, pUniFind identified more peptides in 61.9\% of the modification types (\textbf{Supplementary Fig. S5}), demonstrating its broad applicability and robustness across a wide range of PTMs.

\textbf{Performance evaluation on Astral and timsTOF data with more validation strategies}

\begin{figure}[H]
\centering
\includegraphics[width=1\textwidth]{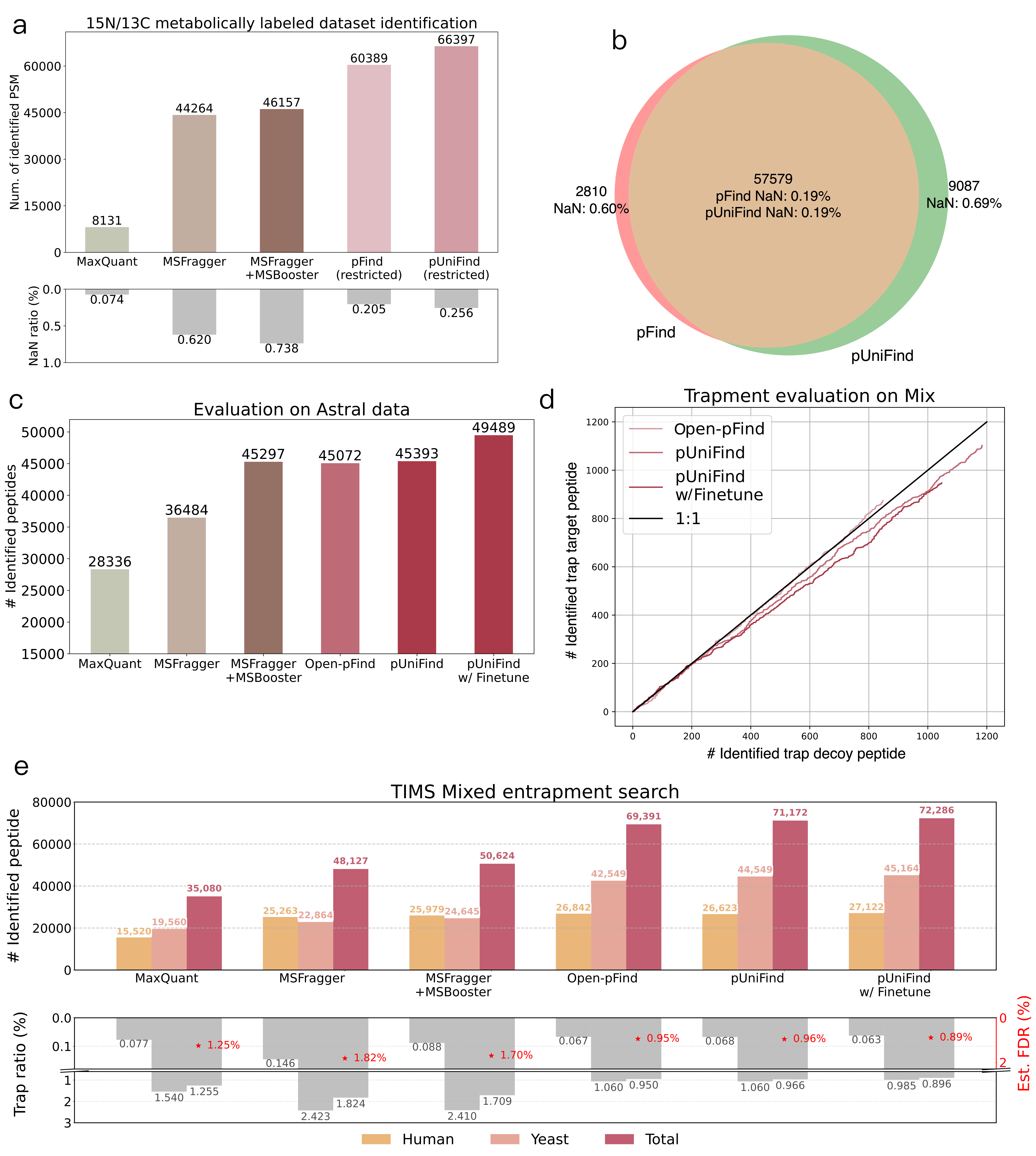}
\caption{\textbf{Further accuracy evaluation of pUniFind and evaluation on trapped ion mobility spectrometry (TIMS) and Astral.} 
\textbf{a}. and \textbf{b}. Evaluation of accuracy on Dong-\textit{E. coli}-QE with $^{15}$N or $^{13}$C labeling method.
\textbf{c}. Evaluation of various software tools on Astral data.
\textbf{d}. Line plot of target and decoy peptides numbers for Open-pFind, pUniFind and pUniFind with finetuning on the TIMS human and yeast mixed data search.
\textbf{e}. Peptide identification of two timsTOF datasets in a mixed search against a combined human and yeast FASTA.
}\label{fig3}
\end{figure}

We further employed more validation strategies and conducted in-depth evaluations on more diverse types of mass spectrometry data. First, we assessed search engine precision using a metabolically labeled \textit{E. coli} dataset (\textbf{Fig. \ref{fig3}a}). Based on the design principle of this dataset, if a PSM for an unlabeled peptide is correct, the corresponding $^{15}$N- or $^{13}$C-labeled precursor ions should typically be detected in MS1 scans, resulting in valid quantification ratios. In other words, when the labeled precursor corresponding to an unlabeled peptide is undetected, leading to Not-a-Number (NaN) quantification ratios, the PSM is more likely to be incorrect. Generally, a lower percentage of NaN ratios indicates higher identification accuracy. Since the pQuant software cannot correctly evaluate quantification results for unexpected modifications, we restricted our analysis to peptides that are not modified or modified with common modifications used in restricted database search, which constitute the majority of all results\cite{chi2018comprehensive}. All search engines were tested against the complete UniProt reviewed FASTA database (released in Oct. 2023, containing 570,830 protein sequences) to enable more competitive benchmarking on a larger database. The evaluation results demonstrate that pUniFind identified 10\% more PSMs than pFind and 43.9\% more than MSFragger with MSBooster, while producing only 0.26\% NaN ratios. This ratio is significantly lower than that of MSFragger with MSBooster and only slightly higher than pFind. Although MaxQuant yielded the lowest percentage of NaN results, its identification count was substantially lower than other engines in such large-scale search—aligning with our previous observations\cite{wang2023universal}. Consistency analysis revealed that pUniFind’s PSMs covered 95.1\% of those identified by Open-pFind, with this overlapping subset exhibiting the lowest NaN ratio of 0.19\% (\textbf{Fig. \ref{fig3}b}). Furthermore, the NaN ratios for uniquely identified peptides were comparable, with 0.60\% for Open-pFind and 0.69\% for pUniFind. These results underscore pUniFind’s superior performance in peptide identification, offering both higher accuracy and more consistent results with lower NaN ratios compared to other commonly used proteomics engines.

We further evaluated pUniFind’s performance on several other types of mass spectrometers, i.e., timsTOF and Astral. 
For this experiment, we first retrained the model by incorporating timsTOF data (\textbf{Supplementary Table S1}) to serve as the default model. A rapid fine-tuning strategy was then applied to adapt the model to both timsTOF and Astral data:
(1) collecting peptides and their highest-scoring spectra identified by Open-pFind (including decoys, to avoid training bias); (2) treating 
best-ranked peptides as positive instances and target peptides ranked 3rd–10th as negatives; (3) fine-tuning for just one epoch. This yielded significant improvements: on Astral data, pUniFind’s performance increased by 9\% after fine-tuning compared to Open-pFind and MSFragger with MSBooster (\textbf{Fig. \ref{fig3}c}). 

For timsTOF data, pUniFind identified 42.7\% more peptides than MSFragger with MSBooster(\textbf{Fig. \ref{fig3}e}).
To rigorously validate the accuracy of our model, we implemented a mixed-species entrapment strategy on timsTOF data.
Briefly, the two datasets with different species, i.e., human and yeast, were co-searched, with each species’ database serving as the entrapment for the other’s MS/MS data. If the deep learning model memorized target peptide features from one species during fine-tuning, it would generate entrapment identifications in the other species. Results indicate that pUniFind—both before and after fine-tuning—retains its identification advantage while achieving substantially lower entrapment ratios compared to all other software, including Open-pFind. Moreover, pUniFind with fine-tuning exhibits the lowest estimated FDR (0.89\%), which is significantly lower than that of MSFragger with MSBooster (1.7\%), a method that represents traditional deep learning feature-based percolator scoring. Furthermore, \textbf{Fig. \ref{fig3}d} demonstrates that our model does not exhibit a preference for assigning higher scores to target peptides compared to decoy peptides. Instead, it displays more conservative scoring, with target peptides receiving slightly lower scores. In summary, comprehensive evaluations across various MS data types and validation strategies confirm pUniFind’s consistent advantage in both identification yield and accuracy. While the large model enhances discriminative power, our rigorous data annotation and training protocols ensure reliable results.

Considering that data analysis methods those for Astral and timsTOF data are less mature than conventional MS data, we speculate that pUniFind could achieve even better results with larger training datasets and more advanced data annotation.

\textbf{Open de novo sequencing workflow}

Open de novo peptide sequencing, which reports results containing any known modifications in a manner similar to open search, is significantly more challenging than traditional de novo sequencing. The larger search space for potential peptide candidates, due to the simultaneous consideration of various post-translational and chemical modifications, substantially increases complexity. Additionally, the lack of large-scale training datasets containing diverse modifications, coupled with the fact that different modifications affect the peptide length to varying degrees, further complicates the accurate prediction of peptides directly from mass spectra.

\begin{figure}[H]
\centering
\includegraphics[width=1\textwidth]{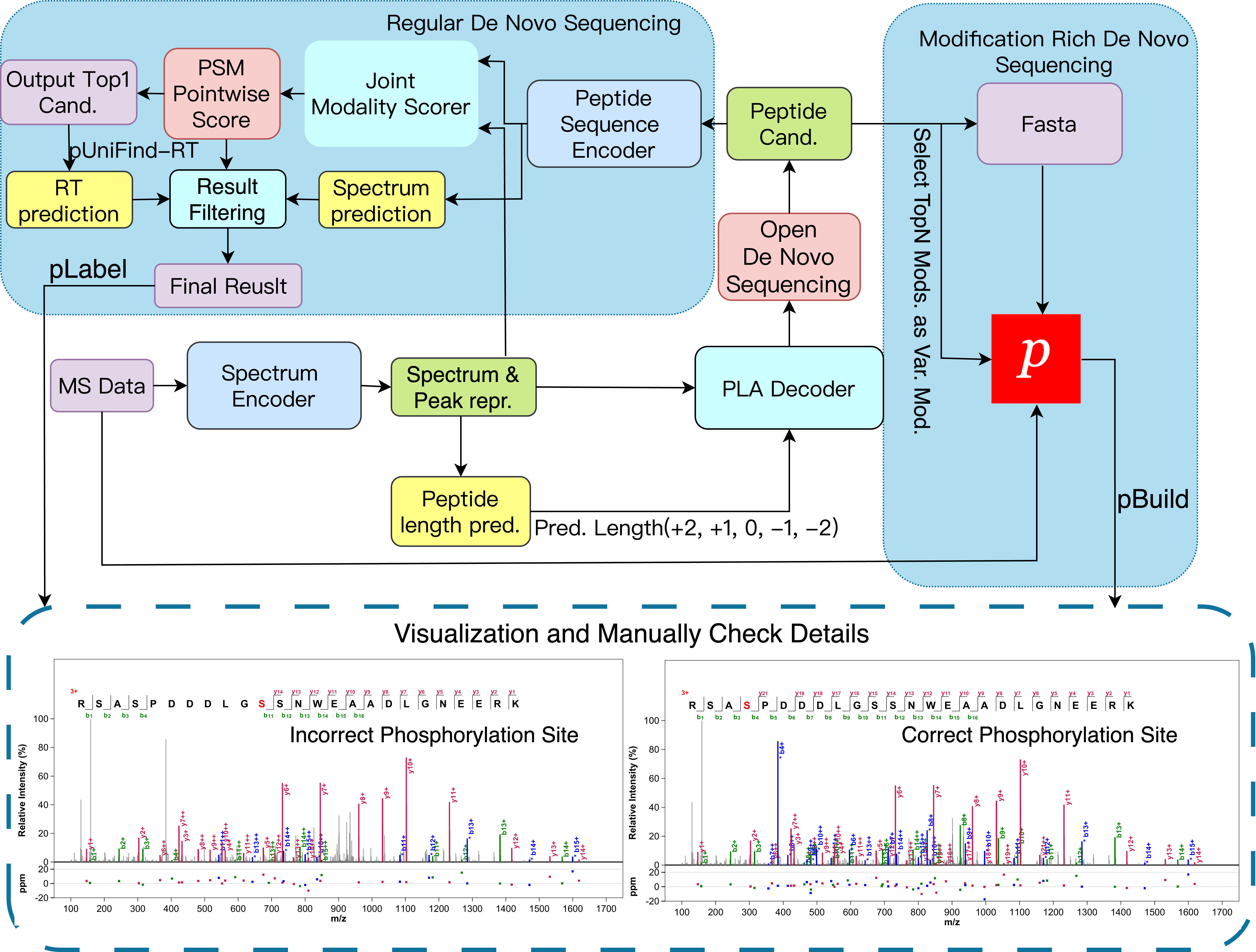}
\caption{\textbf{Workflow of pUniFind in open de novo sequencing.} pUniFind encompasses two workflows—regular de novo and modification-rich de novo—to process various types of data. Since the modification-rich de novo workflow employs pFind3 for result reporting, pBuild can be utilized to view and manually check the results reported. Our software also automatically reports files required by pLabel to visualize results. For example, both the left and right PSM share an identical peptide sequence with the same phosphorylation modification (indicated by the red residue). However, differences in the modification site result in distinct peak matching patterns, with the left PSM exhibiting significantly fewer matched peaks than the right.}\label{fig4}
\end{figure}

Based on pUniFind, we have designed and implemented the first deep learning-based open de novo sequencing pipeline (\textbf{Fig. \ref{fig4}}). First, peptide length prediction serves as a pre-training task. Validation set analysis shows that for over 99\% of spectra, the predicted length deviates from the actual length by no more than 2 amino acids. Then, for each spectrum, the corresponding peptide sequence with its potential modifications is predicted within a peptide length range of $\pm$2 amino acids from the predicted value. In this study, we propose two de novo sequencing workflows: (1) regular de novo sequencing and (2) modification-enriched de novo sequencing. The former is suitable for most conventional datasets with dispersed modification types where each modification type accounts for a small proportion of the total data, while the latter is a more refined workflow for datasets containing modifications that are both abundant and of particular interest to users, such as those enriched in phosphorylation. For regular de novo sequencing, we use the joint modality scorer module implemented in pUniFind for PSM scoring, selecting the highest-scoring peptide as output. For modification-enriched de novo sequencing, since rare modifications typically receive lower scores than regular amino acids in the process of residue prediction, we incorporate the database search engine pFind to rescore the results obtained from de novo sequencing. Specifically, all de novo sequencing results are compiled into a FASTA database file, and the restricted search mode of pFind is used to search this database, with a particular focus on the top four modifications predicted by the deep learning model as well as any additional modifications specified by users. 
This approach is similar to Metanovo\cite{potgieter2023metanovo}, which also incorporates a subsequent database search to refine the de novo sequencing process, but it accounts for a broader range of modifications automatically detected by pUniFind.
In addition, our deep learning model predicts tokenized modification types (e.g., acetylation or formylation) rather than modifications with their sites to make use of training data with other sites that share the same mass. For each spectrum, the highest-scoring peptide identified by pFind serves as the final result. Notably, each spectrum during the de novo sequencing step may be re-identified as a different peptide from other PSMs in the subsequent database search, providing an additional opportunity to refine the initially reported results. 

Finally, the lack of effective quality control methods has limited the application of de novo sequencing. Leveraging our multitask-based training, we developed a deep learning feature-based filtering strategy for de novo sequencing that effectively eliminates unreliable results while maintaining reasonable recall rates. This approach incorporates several deep learning features extracted based on end-to-end scores, predicted spectra, and retention times to filter out unreliable PSMs (see details in Online Methods).

\textbf{Application of pUniFind in open de novo sequencing of diverse modification-rich and regular datasets}

\begin{figure}[H]
\centering
\includegraphics[width=1\textwidth]{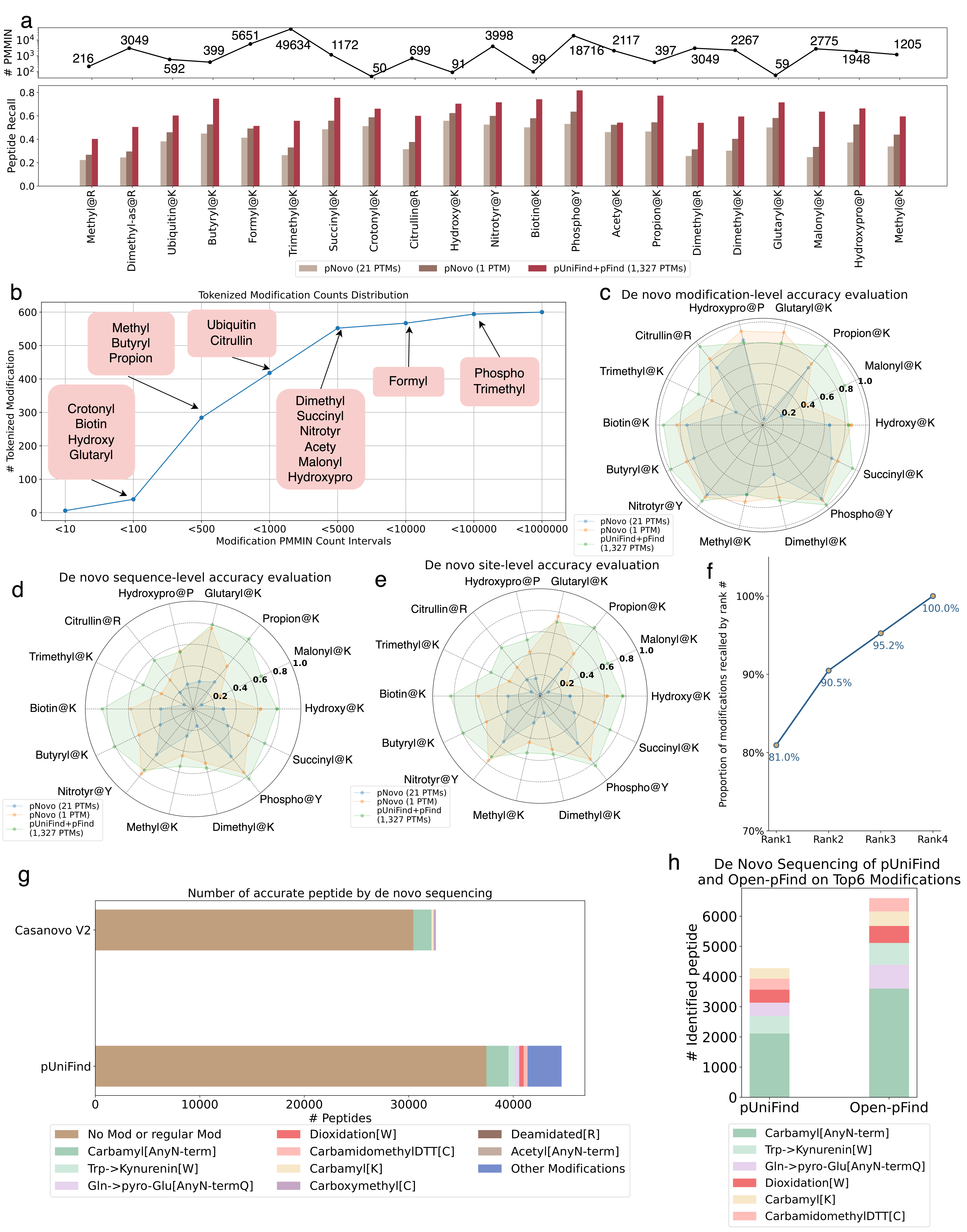}
\caption{\textbf{De Novo sequencing evaluation} 
\textbf{a}. Peptide recall for the 21 PTM datasets and PMMIN count in the training set. 
\textbf{b}. Tokenized modification (disregarding modification sites) count distribution in the training set. For example, counts for Acetyl[AnyN-term] and Acetyl[K] are added up.
\textbf{c}. \textbf{d}. \textbf{e}. More detailed de novo sequencing evaluation of pUniFind and pNovo at various modification levels is presented. Modification level accuracy is defined as follows: for all PSMs with modifications beyond Carbamidomethyl[C], a PSM is considered correct if pUniFind predicts the exact same modifications as those identified by Open-pFind. Similarly, site-level accuracy is assessed for all PSMs with modifications beyond Carbamidomethyl[C]. In this case, a PSM is deemed correct if both the modifications and their sites, as predicted by pUniFind, match exactly with those determined by Open-pFind.
\textbf{f}. Proportion of the 21 PTMs for which the correct modification was recalled by pUniFind within the top four ranked modifications (ranks 1 to 4). For each PTM dataset, modifications were ranked based on the total number of peptide candidates (carrying each modification) across all spectra.
\textbf{g}. Regular open de novo pipeline peptide level evaluation on \textit{S. cerevisiae} data. Regular modifications include Carbamidomethyl[C], Oxidation[M], Acetyl[ProteinN-term].
\textbf{h}. Peptide counts for each of the top 6 modifications for pUniFind and Open-pFind.
}\label{fig5}
\end{figure}

We analyzed the 21PTM dataset, consisting of 21 sub-datasets with diverse PTMs, to evaluate pUniFind's capability in open de novo peptide sequencing. 
Open-pNovo was excluded from benchmarking due to its consistently low performance in preliminary evaluations. Specifically, it exhibited a peptide-level recall below 10\% across 21PTM datasets, likely reflecting inherent limitations in its underlying machine learning framework. As an alternative, we included pNovo in our evaluation, using two settings: one with all 21 PTMs collectively included, and another with the corresponding modification specified for each individual dataset.
On this dataset, pUniFind combined with pFind achieved an average peptide-level recall of 63.8\% (\textbf{Fig. \ref{fig5}a}). Despite operating in a 300-fold expanded search space \cite{yang2017open}, pUniFind performed comparably to conventional de novo sequencing methods (an average peptide-level recall of 64\%\cite{yilmaz2024sequence}) while yielding a a 60\% improvement in performance compared to pNovo across all 21 sub-datasets.
Notably, pUniFind performs de novo sequencing by comprehensively considering over 1,300 modification types without any prior knowledge of modifications. In contrast, even when pNovo is individually configured with the appropriate modification for each sub-dataset, its average peptide-level recall remains at 47.6\%, markedly lower than that of pUniFind. Moreover, when pNovo is set to consider all 21 PTMs, its performance further declines (average recall: 39.8\%) due to the substantially expanded search space. 
Remarkably, in each of the 21 PTM datasets, the target PTM ranked within the top four by number of identified PSMs, and was the most frequently identified modification in 81\% of the datasets \textbf{Supplementary Table S2}. 
The relatively low ranking (fourth) of Crotonyl[K] modification can be attributed to the fact that the mass of Crotonyl[K] precisely matches the combined masses of residues Proline (P) and Valine (V) or Valine and Proline (V+P). This observation underscores the necessity of incorporating peptide candidates with varying lengths into the prediction process to properly account for such mass coincidences.

We further assessed the accuracy of pNovo and the modification-enriched workflow based on pUniFind across three levels: modification, sequence, and site (\textbf{Fig. \ref{fig5}c-e}). The results reveal that pUniFind achieves over 90\% accuracy at the modification level for most datasets. If the search space of pNovo was limited to dataset-specific PTMs, it outperformed pUniFind in a few cases, e.g., in Hydroxyproline datasets. However, as the number of modifications increased, pNovo’s accuracy declined significantly across all levels evaluated. At the more stringent peptide sequence and site levels, the modification-enriched workflow based on pUniFind consistently outperformed pNovo, demonstrating that pUniFind provided more accurate results for both standard and modified amino acids in a broader sequencing space.

Additionally,  for lysine-targeted modifications such as crotonylation, hydroxylation, biotinylation, and glutarylation—each having fewer than 100 PMMINs (Peptide-spectrum matches with the same peptide sequence, Modification type, Modification site, Instrument, and Normalized collision energy, NCE) in the training set—pUniFind still achieves a peptide recall rate greater than 60\% (\textbf{Fig. \ref{fig5}b}). Generally, the recall rates improve as the number of PMMINs increases. Among the 600 tokenized modifications in the training set, 554 (92.3\%) have more than 100 PMMINs. Thus, it is reasonable to infer that pUniFind performs well even for rare modifications.

Furthermore, we applied the open de novo sequencing workflow to a regular \textit{S. cerevisiae} dataset. Like Open-pFind, which significantly increased the identification rate for database search, our study demonstrates for the first time that the open de novo sequencing workflow based on pUniFind identified 40.2\% more peptides compared with Casanovo V2 (\textbf{Fig. \ref{fig5}g}). Additionally, the six most abundant modifications identified by pUniFind in the open de novo sequencing approach were identical to those reported by Open-pFind (\textbf{Fig. \ref{fig5}h}), thereby confirming the robustness of the modification discovery process. However, slight differences were observed in their ranking, likely reflecting variable detection performance across distinct modifications. This exceptional performance is consistent with additional evaluation results obtained from nine species datasets using labels annotated by other search engines (\textbf{Supplementary Fig. S6–S7}).

\subsection*{Enhancing mass spectra interpretation for challenging cases}
\textbf{Challenges of large-scale database searches}

\begin{figure}[H]
\centering
\includegraphics[width=1\textwidth]{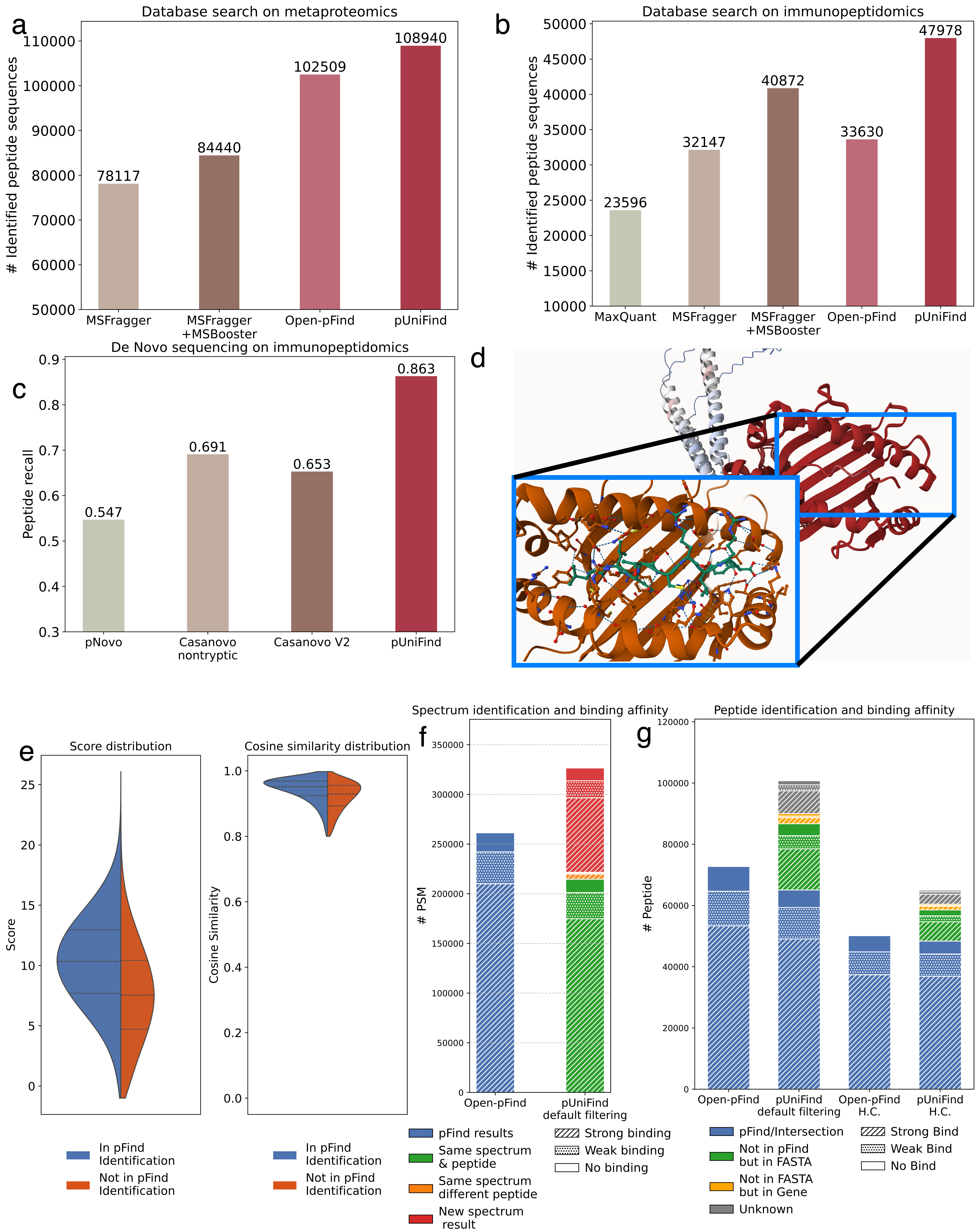}
\caption{\textbf{Application of pUniFind on various hard cases.} 
\textbf{a}. Evaluation of peptide identification via database search in the metaproteomics dataset.
\textbf{b}. Evaluation of peptide identification via database search in the immunopeptidomics dataset.
\textbf{c}. Evaluation of de novo sequencing performance in the immunopeptidomics dataset.
\textbf{d}. Structure of the HLA peptide ``LLDTCRLRY" predicted by pUniFind, which is absent from the human protein database but present in the gene database. Structures of the peptide binding to HLA-A0101 were predicted by AlphaFold3 using five random seeds, with the resulting structures aligned. The main image colors residues by pLDDT confidence scores, where red indicates high reliability and white denotes low-confidence regions. The inset shows the interaction between ``LLDTCRLRY" and HLA-A0101, with protein chains colored by chain ID. 
\textbf{e}. Distribution of scores and cosine similarity for filtered de novo sequencing results, comparing identified PSMs versus non-identified ones by pFind.
\textbf{f, g}. Evaluation of filtering strategies at the PSM and peptide levels using several metrics, including the consistency with Open-pFind, the presence in sequence databases, and the binding affinity. H.C. stands for high confidence, indicating no missing fragment ion sites.}

\label{fig6}
\end{figure}

In scenarios with ultra-large search spaces—where the number of candidate peptides per spectrum increases dramatically—accurate scoring becomes significantly more challenging. To assess pUniFind’s robustness under these demanding conditions, we systematically evaluated its performance improvements in peptide identification across complex applications such as metaproteomics and immunopeptidomics. We first reanalyzed the metaproteomics data published by Gavin et al.\cite{gavin2018intestinal} and found that pUniFind still identified 6.3\% and 29.0\% more peptides than Open-pFind and MSFragger with MSBooster, respectively (\textbf{Fig. \ref{fig6}a}). 
The overlap in peptide-level identification between pUniFind and other search engines is presented in \textbf{Supplementary Fig. 8}, demonstrating that pUniFind achieves high consistency with other search engines, particularly with Open-pFind. cosine similarities between identified and unidentified precursors are compared in \textbf{Supplementary Fig. 9}. Precursors not identified by Open-pFind still exhibit high similarity scores (median: 0.93), although slightly lower than those identified by both engines (median: 0.95). 
This difference is expected, as precursors missed by Open-pFind are likely to be of lower spectral quality.
These results highlight pUniFind’s consistent advantage in large-scale applications, where the complexity and database size of metaproteomic data significantly exceed those of conventional proteomics analyses. Immunopeptidomics data presents an even greater challenge, as the search space expands tens of folds compared to conventional datasets due to the inclusion of non-specific digestion. For the immunopeptidomics data from Michal Bassani-Sternberg et al.\cite{bassani2016direct}, MSFragger with MSBooster—by leveraging deep learning features such as predicted spectra and retention time—outperformed Open-pFind (40,872 vs. 33,630, \textbf{Fig. \ref{fig6}b}). However, pUniFind still identified the highest number of peptides, surpassing MSFragger with MSBooster and Open-pFind by 17.4\% and 42.6\%, respectively. 
It also covered 91.9\% of peptide identifications made by MSFragger with MSBooster, and 97.5\% of those identified by Open-pFind, indicating high consistency and reliability (\textbf{Supplementary Fig. 10}). 
Furthermore, the median cosine similarity of precursors uniquely identified by pUniFind (0.95) was identical to that of those also identified by Open-pFind, supporting the reliability of pUniFind-specific identifications (\textbf{Supplementary Fig. 11}).
Together, these results highlight pUniFind’s robust performance advantage in handling non-specifically digested peptides.



\textbf{Application of de novo sequencing and result filtering in immunopeptidomics}

Conventional evaluation of de novo sequencing models relies on PSMs identified through database search engines. However, this validation framework is inherently limited, because it depends on external databases, making it unsuitable for assessing de novo sequencing in real-world applications where database-independent identification is crucial. 
Our study employed a comprehensive evaluation protocol comprising three main steps: (1) performing de novo sequencing and database search on all experimental MS/MS spectra; (2) applying deep learning-based spectral filtering; and (3) conducting a comparative analysis of search engine and de novo sequencing results across both protein and genomic databases, including the identification of novel peptides absent from existing references.

As shown in \textbf{Fig. \ref{fig6}e}, the distribution of PSM identifications within and outside Open-pFind is broadly similar, although PSMs uniquely identified pUniFind shows slightly lower scores and cosine similarities. This is expected, as spectra identified by Open-pFind are generally of higher quality.
\textbf{Fig. \ref{fig6}f} demonstrates that the ratio of strong and weak binding peptide identifications for both pUniFind and Open-pFind is very high at the PSM level, with only 3.2\% of spectra identified by both methods corresponding to different peptides. At the peptide level, pUniFind recalled 89.8\% of peptides identified by Open-pFind (\textbf{Fig. \ref{fig6}g}). Meanwhile, 89.3\% of peptides are found in the human proteome or genome-translated database, with 3.3\% (3,303) of peptides existing only in the latter. Furthermore, the ratio of peptides found in these two databases increased to 93\% when the corresponding PSMs with missing fragment ions are filtered out, which are referred to as the high-confidence results (\textbf{Figs. \ref{fig6}f-g}). These findings indicate that our filtering strategy effectively improves result reliability and increases the proportion of identifications matching existing databases. \textbf{Fig \ref{fig6}d} shows one case of an HLA peptide which is not in human protein database but in human gene database identified by pUniFind. In the end we found 1,891 peptides in human gene database but not in human protein database with no missing fragment ions satisfying our default filtering standard. The structural model, colored according to the pLDDT predicted by AlphaFold3, indicates high confidence in both the binding sites and the HLA-bound peptide conformations. The predicted inter-chain TM-score (ipTM) between two chains is 0.93. We also predicted the structure for each allele using HLA peptides sequenced by pUniFind that did not exhibit any missing ion fragment sites. All ipTM scores predicted by AlphaFold3 are above 0.90. Notably, these peptides were in genome-translated database rather than the proteome one, as illustrated in \textbf{Supplementary Data 1}.

To complement these database-independent evaluations, we also assessed the performance of pUniFind using conventional database-based benchmarks. pUniFind achieved a peptide recall of 86.3\%, significantly higher than the number reported by the non-tryptic version of Casanovo V2 (69.1\% in \textbf{Fig. \ref{fig6}c}). To avoid data leakage, we deleted all spectra with peptide label in our training set or Casanovo V2 training set. This strong performance underscores pUniFind’s potential for identifying mutated peptides in immunopeptidomics applications. Since the accurate interpretation of non-tryptic data is also essential for tasks such as antibody sequencing, we assessed pUniFind’s generalizability across various enzymes (\textbf{Supplementary Fig. S12}). The results demonstrate that pUniFind consistently outperforms Casanovo V2, highlighting its robustness in non-tryptic scenarios.

\section*{Discussion}\label{sec4}

In this paper, we present pUniFind, a unified multimodal pre-trained model that, for the first time, integrates the refinement of open database search results and deep learning-based open de novo sequencing into a single framework. We employed a cross-modality pre-training strategy—namely de novo sequencing pre-training for the spectrum modality and spectrum prediction for the peptide modality—which enables the deep learning model to better understand spectra and peptides. Our results demonstrate that end-to-end scoring can outperform and potentially replace traditional feature-based scoring methods, thereby fundamentally improving the scoring framework in computational proteomics. pUniFind significantly improves performance on challenging tasks such as immunopeptidomics and metaproteomics. Furthermore, for the first time, we achieved deep learning-based open de novo sequencing, enabling de novo sequencing performance on modified peptides comparable to, or even surpassing, that of unmodified peptides.

It is important to note that pUniFind supports multi-GPU inference to accelerate processing. For example, with a single 4090 GPU, pUniFind takes approximately 65 minutes to analyze the 24 raw files in the \textit{V. mungo} data used in this study (excluding data preprocessing and report generation). However, with eight 4090 GPUs, the inference time is reduced to just 10 minutes, showcasing its high scalability. Overall, pUniFind completes data processing, inference, and report generation in about 15 minutes. In contrast, Open-pFind requires approximately 5 hours to complete an open search, covering a significantly larger search space than a restricted search. Additionally, benefiting from large-scale parallel computation, Tesorai search completes its restricted search in 90 minutes, while DDA-BERT requires more than a day to perform a restricted search on a 3090 GPU (note that DDA-BERT only supports bf16 precision and therefore could not utilize the 4090 GPU, which lacks bf16 support).

Currently, pUniFind does not incorporate retention time into the database search task due to significant variations across different instruments and laboratories, indicating potential for further improvement. However, we plan to integrate retention time in future iterations of our development. Additionally, although our model outperforms state-of-the-art search engines for data-dependent acquisition (DDA) across various instruments and datasets, it has not yet been able to fully leverage the characteristics of data-independent acquisition (DIA) data to enhance its analysis. We believe our methodology could be adapted to refine the scoring framework for DIA data. Future research will focus on constructing large-scale DIA datasets annotated via open search methodologies and extending our unified interpretation framework to DIA data analysis.


\backmatter

\clearpage

\bibliography{sn-bibliography}
\clearpage

\section*{Data availability}\label{dataa}
All raw data used for pre-training and evaluation in this study were publicly available. All meticulously annotated and clustered training data—including data that has been thoroughly searched, analyzed, and filtered—will be made publicly available upon acceptance of our paper. All results generated using pUniFind, pFind, Fragpipe, Tesorai Search, MaxQuant, and DDA-BERT will be made available on the PRIDE platform.

\section*{Code availability}\label{codea}
pUniFind software is available at \href{https://github.com/pFindStudio/pUniFind}{https://github.com/pFindStudio/pUniFind}, including the Windows version executable application, Linux deployment source code, web server link, and user guide. Our source code will be made publicly available upon the acceptance of our paper.

\section*{Acknowledgments}\label{ack}
We would like to thank Guolin Ke from DP Technology for providing part of the computational resources during the research. We are also grateful to Peter Cimermancic from Tesorai, Jinqiu Xiao from Jürgen Cox’s lab, and the Nesvizhskii lab for their valuable suggestions regarding software usage during the evaluation. We thank the iMetalab team for providing the FASTA file for evaluation. We appreciate the authors of Casanovo for their helpful explanations regarding the tool’s usage and data download procedures. We gratefully acknowledge the support of the National Natural Science Foundation of China (Grant No. 32471501).

\section*{Author contributions}\label{aut_con}
Z.-J.L. conducted the project by assembling training data, training the model, and performing data evaluations. 
Z.-W.J. carried out part of the evaluation and developed the Bohrium web server and Windows version. 
C.H., W.H. and Z.-W.J. supervised the project. 
E.-W.N. provided valuable overall suggestions.
W.H. and Z.-W.J. provided suggestions about HLA binding affinity analysis.
M.-P.Z., W.-K.F., and P.-Y.P. offered valuable suggestions and provided essential scripts. 
L.-Y.M., L.-S.Q., and J.-X.H. contributed insights regarding model design. 
C.-R.F. provided critical guidance and assistance with the TIMS and Astral data processes and analyses. 
Z.X. and D.-J.X. supplied a portion of the pre-training data.
L.-Y.C. helped analyze evaluation data.
Computational resources were provided by W.H. and L.-S.Q. 
The manuscript was drafted by Z.-J.L., C.H., W.H. and Z.-W.J.
All authors reviewed and revised the manuscript.

\section*{Competing interests}\label{competing}
The authors declare no competing interests.

\end{document}